\documentclass{article}

\usepackage[dblblindworkshop,final]{neurips_2026}
\workshoptitle{Verification in the Age of AI Scientists}

\makeatletter
\renewcommand{\@noticestring}{Submitted to the AI for Science workshop (NeurIPS 2026).}
\makeatother

\usepackage[utf8]{inputenc}
\usepackage[T1]{fontenc}
\usepackage{hyperref}
\hypersetup{hidelinks}
\usepackage{url}
\usepackage{booktabs}
\usepackage{amsmath,amsfonts,amssymb}
\usepackage{graphicx}
\usepackage{microtype}
\usepackage{xcolor}
\usepackage{array}
\usepackage{tabularx}

\graphicspath{{./}}

\title{Can AI Scientists Change Their Minds?\\
Prior--Evidence Conflict in Synthetic Universes}
\author{%
  Kargi Chauhan\\
  University of California, Santa Cruz\\
  \texttt{kchauha3@ucsc.edu}
}

\newcommand{\systemname}{\textsc{Synthetic Universes}}
\newcommand{\nrmse}{\operatorname{NRMSE}}
\newcommand{\emargin}{\mathcal{M}}

\begin{document}
\maketitle

\begin{abstract}
Can a scientific agent distinguish a law it inferred from evidence from one it merely recognizes? We introduce \systemname{}, a controlled benchmark that pairs canonical \emph{famous worlds} with matched \emph{twisted twins} governed by nearby noncanonical mechanisms. We evaluate each reported law twice: by executing it on held-out continuations and transfer settings, and by independently checking whether it recovers the generating mechanism. In the current checkpoint of a pre-specified 60-cell study, 22 trials were graded and one additional run ended in infrastructure failure. Among 20 twin trials, 8 pass predictive verification while 5 recover the generator. The dissociation is bidirectional: six parsable outputs predict successfully while missing the mechanism, whereas three recover the mechanism but fail predictive rollout. Drag exhibits the first pattern (5/5 predictive pass, 1/5 mechanism recovery); Gravity exhibits the second (1/5 predictive pass, 4/5 mechanism recovery). Because matched famous controls, the corrected identifiability sweep, and the Evidence Ladder remain incomplete, we do not claim a confirmatory causal prior-conflict effect. Instead, the completed runs establish a narrower verification result: \emph{predictive adequacy and mechanism recovery are distinct scientific claims and require distinct tests}.
\end{abstract}

\section{Introduction}

An AI scientist can give the right answer for the wrong reason. Suppose a model observes an orbital trajectory and returns $F\propto r^{-2}$. It may have inferred the governing law from the observations; it may instead have recognized the setting and retrieved Newtonian gravity from pretraining. On a canonical task those explanations are observationally confounded.

This ambiguity matters as automated-science systems move from hypothesis generation to experiment execution and discovery \citep{langley1987scientific,waltz2009automating,lu2026aiscientist,gottweis2025coscientist,novikov2025alphaevolve}. Systems such as FunSearch and AlphaTensor illustrate the value of objective evaluators: open-ended generation becomes useful when outputs can be tested externally \citep{romeraparedes2024funsearch,fawzi2022alphatensor}. As scientific generation scales, verification becomes a bottleneck \citep{cornelio2026verification,mossel2025refutability}. This question sits inside a broader shift from AI as a predictor to AI as a participant in scientific workflows, spanning protein structure prediction, materials discovery, and autonomous chemistry \citep{jumper2021alphafold,merchant2023materials,boiko2023autonomous}.

Instance freshness alone is insufficient. A newly generated trajectory can still have a centuries-old answer strongly represented in pretraining. \systemname{} therefore turns \emph{prior--evidence agreement} into an experimental variable. Every domain contains a canonical \emph{famous world} and a nearby \emph{twisted twin}. The interface stays fixed while the generating law changes. In Gravity, for example, $r^{-2}$ is paired with $r^{-2.3}$. The familiar hypothesis remains plausible, but it is wrong. Fresh or dynamically generated evaluations reduce reuse of particular benchmark items, but they do not remove a model's prior over canonical scientific answers \citep{white2025livebench,li2024latesteval,xu2024contamination,sainz2023contamination}.

A second distinction is equally important: predicting unseen data does not necessarily identify the true mechanism. Two laws may extrapolate similarly over a limited intervention. We therefore score \emph{predictive pass} and \emph{mechanism recovery} independently.

We use ``change their minds'' operationally, not anthropomorphically: the question is whether the agent's final executable law departs from the canonical family when the supplied observations support a noncanonical alternative.

A useful decision-theoretic idealization makes the source of the conflict explicit. Let $\mathcal F$ be a hypothesis space, $P(f)$ a structural prior, and $\mathcal L(D\mid f)$ the likelihood of observations $D$ under law $f$. For a canonical candidate $f_F$ and a twisted alternative $f_T$,
\begin{equation}
\log\frac{P(f_T\mid D)}{P(f_F\mid D)}
=
\log\frac{\mathcal L(D\mid f_T)}{\mathcal L(D\mid f_F)}
+
\log\frac{P(f_T)}{P(f_F)}.
\label{eq:posterior_odds}
\end{equation}
Evidence overcomes a canonical prior when the log-likelihood ratio exceeds the opposing log-prior odds. We do not claim access to an LLM's internal $P(f)$; \systemname{} instead manipulates whether familiar knowledge agrees with the evidence and measures the law the agent ultimately commits to.

\begin{figure}[t]
    \centering
    \includegraphics[width=0.84\linewidth]{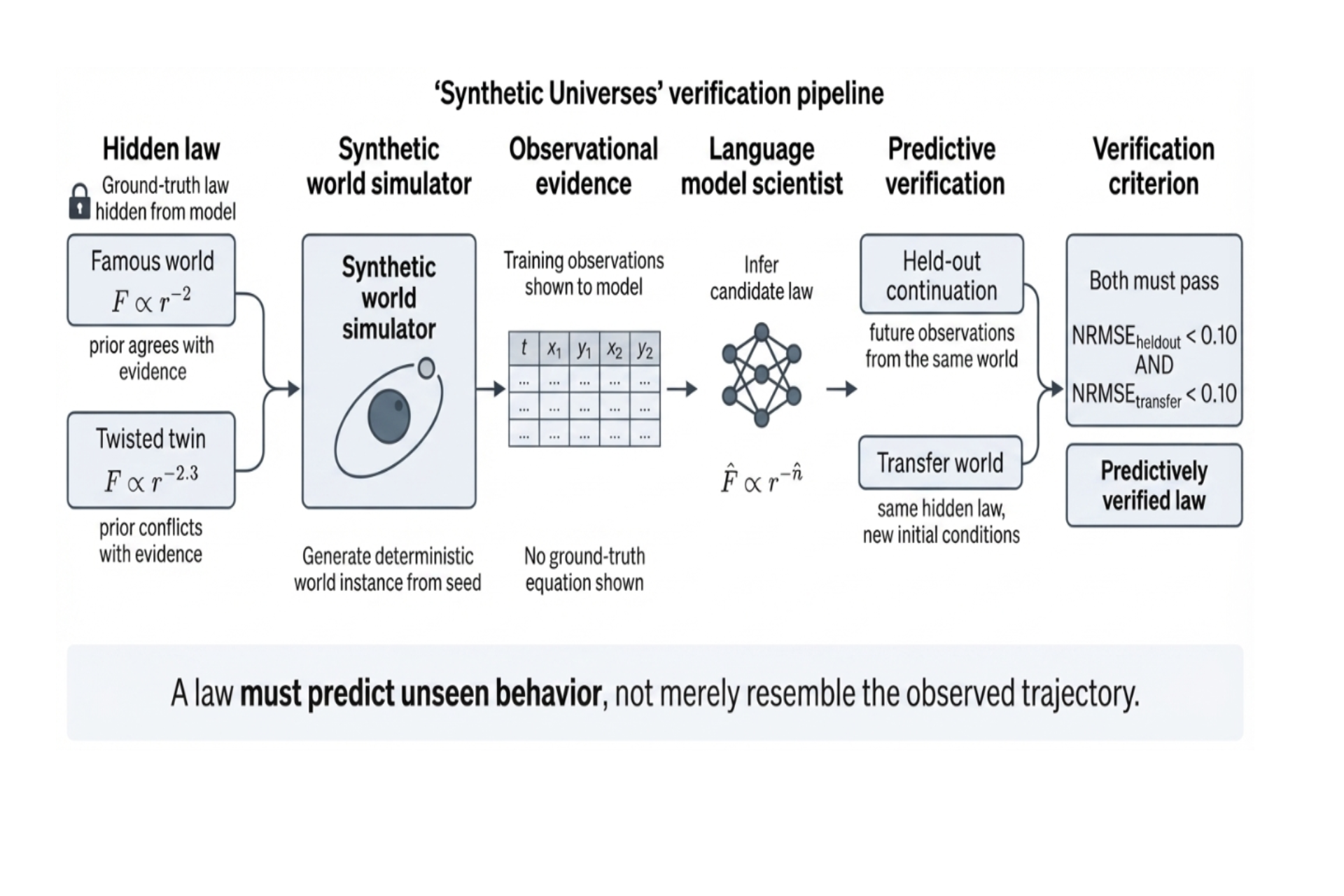}
    \caption{\textbf{\systemname{} separates predictive verification from mechanism recovery.}
    The agent sees observations but not the hidden law. Its reported equation is executed on an unseen continuation and a transfer world, then independently checked against the generating mechanism. A predictive pass therefore certifies performance on the tested interventions, not mechanism discovery.}
    \label{fig:pipeline}
\end{figure}

Our benchmark asks three questions: (i) does a scientific agent behave differently when familiar knowledge conflicts with the observations; (ii) when it fails, did the experiment contain enough information to discriminate the alternative; and (iii) what additional evidence would be sufficient to move the agent away from a familiar explanation? The present submission reports the benchmark, the frozen verification taxonomy, and the completed subset of the pre-specified replication. It does \emph{not} treat the unfinished matrix or planned interventions as completed confirmatory evidence.

\section{Related Work}

\paragraph{AI scientists and verification.}
The AI Scientist automates substantial portions of the research loop \citep{lu2026aiscientist}; Google's AI Co-Scientist generates and refines scientific hypotheses \citep{gottweis2025coscientist}; AlphaEvolve combines language-model generation with executable evaluators \citep{novikov2025alphaevolve}; Gemini case studies explore code-supported scientific problem solving \citep{woodruff2026gemini}; and multi-agent systems increasingly automate scientific work \citep{ghareeb2026multiagent}. Recent work explicitly emphasizes that persuasive scientific narratives or even apparently correct conclusions need not establish the claimed mechanism \citep{cornelio2026verification,eulig2026wrongmechanism,bai2026storyscience}. Critic/falsifier loops can provide additional scrutiny \citep{bansal2026pcf}, but here we begin with a single-agent setting to isolate evidence updating.

\paragraph{Scientific discovery beyond memorization.}
DiscoveryBench, ScienceAgentBench, ResearchBench, and related evaluations move beyond static science QA toward research workflows \citep{majumder2025discoverybench,chen2025scienceagentbench,liu2026researchbench,song2025sde}. LLM-SR explicitly combines language-model priors with programmatic and evolutionary equation search \citep{shojaee2025llmsr}; LLM-SRBench evaluates equation discovery under transformations and synthetic problems designed to reduce trivial memorization \citep{shojaee2025llmsrbench}; and NewtonBench studies altered physical laws in interactive simulators \citep{zheng2025newtonbench}. \systemname{} isolates a complementary variable: whether a familiar scientific prior agrees or conflicts with otherwise matched evidence. The present checkpoint does not yet include a prior-free symbolic baseline or cross-model replication, so we treat any attribution specifically to language-model pretraining priors as a hypothesis to be tested rather than a settled causal conclusion.

\paragraph{Equation discovery and priors.}
Symbolic regression and sparse identification recover compact laws directly from data \citep{schmidt2009natural,brunton2016sindy,udrescu2020feynman,udrescu2020feynman2,cranmer2023pysr}. Language agents bring an additional resource in the form of strong pretrained scientific priors, which can accelerate search when correct but can also act as an attractor when deliberately contradicted. Counterfactual reasoning studies similarly show competition between supplied rules and world knowledge \citep{li2023counterfactual,yamin2026worldknowledge}.

\section{Synthetic Universes}

\subsection{Matched worlds}

A task consists of simulator $S_\theta$, observation operator $O$, and hidden law parameters $\theta$. For each family we construct
\begin{align}
D_F &\sim O(S_{\theta_F}), & \theta_F &\in\Theta_{\mathrm{canonical}},\\
D_T &\sim O(S_{\theta_T}), & \theta_T &\notin\Theta_{\mathrm{canonical}}.
\end{align}
The observation schema and broad task remain fixed:
\[
\text{famous: prior agrees with evidence},
\qquad
\text{twin: prior conflicts with evidence}.
\]
The prompt uses neutral variable names and never labels a task ``gravity,'' ``spring,'' or ``twisted.''

\begin{table}[t]
\caption{\textbf{Six benchmark families.} Continuous families are written using a common exponent $p$. The dagger marks structural families retained for transparent run accounting but not used for strong mechanistic claims until their generator and non-degeneracy audit is complete.}
\label{tab:worlds}
\centering
\small
\begin{tabular}{lll}
\toprule
Family & Famous mechanism & Twisted mechanism\\
\midrule
Spring & $-k\,\mathrm{sgn}(x)|x|^{p}$, $p=1$ &
$-k\,\mathrm{sgn}(x)|x|^{p}$, $p=1.3$\\
Gravity & $F_r\propto r^{-p}$, $p=2$ &
$F_r\propto r^{-p}$, $p=2.3$\\
Drag & $-k\|\mathbf v\|^{p-1}\mathbf v$, $p=1$ &
$-k\|\mathbf v\|^{p-1}\mathbf v$, $p=1.6$\\
Pendulum & $-\omega^2\sin\theta$ &
$-\omega^2\sin\theta(1+0.4|\theta|)$\\
Conservation$^\dagger$ & audited target invariant class & generator under audit\\
Coupling$^\dagger$ & audited pair-interaction class & generator under audit\\
\bottomrule
\end{tabular}
\end{table}

The twists deliberately span two kinds of scientific revision. Spring, Gravity, and Drag are continuous deformations of familiar exponents, whereas Pendulum changes the functional response itself. Conservation and Coupling were intended to test structural equivalence classes, but the current generator audit found an unintended energy-like invariant in Conservation and Coupling remains too sparse to certify its intended contrast. We therefore do not insert intended design equations as if they were audited ground truth. Appendix~\ref{app:structural} states the non-degeneracy checks required before those families return to the headline mechanistic analysis. These cases need not have equal difficulty, so we report family-level behavior rather than treating the six families as exchangeable replicates.

All hidden parameters, trajectories, observation tables, and transfer instances are generated specifically for evaluation. We therefore claim \emph{instance-level novelty}: the benchmark is contamination-resistant, not contamination-proof.

\subsection{Execution-grounded predictive verification}

For trajectory worlds, let $Y_H$ denote a held-out continuation and $Y_T$ a transfer trajectory under new initial conditions. A proposed law $\hat f$ is executed and scored with
\begin{equation}
\nrmse(\hat Y,Y)=
\frac{\sqrt{\frac{1}{N}\sum_{i=1}^{N}\|\hat{\mathbf y}_i-\mathbf y_i\|_2^2}}{s},
\qquad
\hat{\mathbf y}_i,\mathbf y_i\in\mathbb R^d,
\end{equation}
where $s>0$ is a candidate-independent normalization fixed by the generator before any agent output is scored; the exact per-family computation and value are part of the released generator metadata. The same frozen $0.10$ rule is used for the headline label across families so that outcome definitions do not change after inspecting results. Because long-horizon orbital and oscillatory systems can accumulate phase error while dissipative systems can contract, we additionally pre-specify derivative-field, calibrated-rollout, and geometry-aware sensitivity diagnostics in Appendix~\ref{app:dynamics}; these diagnostics do not retroactively alter the headline label. A predictive pass requires
\begin{equation}
\nrmse_{\mathrm{holdout}}<0.10
\quad\land\quad
\nrmse_{\mathrm{transfer}}<0.10.
\label{eq:verify}
\end{equation}
Event-based conservation tasks use the same principle with normalized invariant residuals on held-out and independently generated transfer events.

\subsection{Mechanism recovery}

Predictive success is not mechanism identification. The frozen mechanism checker first parses the reported law into an executable representation and compares its functional form against the target family on a fixed evaluation domain. Algebraically equivalent forms are normalized before comparison, and irrelevant global scaling of conservation quantities is quotiented out. Operationally,
\begin{equation}
\mathrm{Mech}(\hat f)=
\mathbf 1\!\left[
\mathrm{parsed}(\hat f)\land
d_{\mathrm{func}}(\hat f,f^\star)\le\tau_{\mathrm{func}}\land
d_{\mathrm{shape}}(\hat f,f^\star)\le\tau_{\mathrm{family}}
\right],
\label{eq:mechanism}
\end{equation}
with structural families replacing the shape term by a functional-equivalence test. The executable functional-residual gate is $\tau_{\mathrm{func}}=0.05$ where applicable; family-specific shape gates are frozen in the released checker and are applied uniformly, independently of predictive pass. For future structural-family inclusion, equivalence is not sufficient by itself: the generator must also pass the non-degeneracy audit in Appendix~\ref{app:structural}. Appendix~\ref{app:mechanism} gives the equivalence logic and audited examples.

Every completed trial therefore lies in one of four cells:
\begin{center}
\small
\begin{tabular}{lcc}
\toprule
& Mechanism recovered & Mechanism missed\\
\midrule
Predictive pass & verified recovery & predictive alternative\\
Predictive fail & unstable recovery & failed discovery\\
\bottomrule
\end{tabular}
\end{center}

\subsection{Identifiability and evidence interventions}

A failed twin is not evidence of prior interference if the observations themselves cannot distinguish the noncanonical mechanism. For trajectory data $D=\{(t_i,\mathbf y_i)\}_{i=1}^{N}$, define the candidate-independent training loss
\begin{equation}
E(f;D)=\frac{1}{N s^2}\sum_{i=1}^{N}
\|\hat{\mathbf y}_{f}(t_i)-\mathbf y_i\|_2^2,
\label{eq:identloss}
\end{equation}
where $\hat{\mathbf y}_{f}$ is obtained by executing $f$ from the task's fixed initial state with the frozen numerical solver. Let
\[
E_F(D)=\min_{f\in\mathcal F_F}E(f;D),
\qquad
E_T(D)=\min_{f\in\mathcal F_T}E(f;D).
\]
For the continuous deformations, $\mathcal F_F$ is nested inside the more flexible twin-aware family $\mathcal F_T$, so an unpenalized residual ratio is optimistically biased toward $\mathcal F_T$. The corrected control therefore uses a complexity-adjusted Gaussian-residual score,
\begin{equation}
\mathrm{BIC}_h
=
N\ln\!\bigl(E_h(D)+\epsilon\bigr)+k_h\ln N,
\qquad
\emargin_{\mathrm{BIC}}(D)
=
\mathrm{BIC}_F-\mathrm{BIC}_T,
\label{eq:bicmargin}
\end{equation}
where $k_h$ is the number of fitted parameters. Positive values favor the twin-aware family after accounting for its additional degrees of freedom. As a robustness check, the same families should also be compared on a held-out partition of the visible observations. The fitted twin-aware model must then pass the same external holdout and transfer verifier before a twin is called oracle-identifiable. This remains an \emph{oracle parametric upper bound}, not a prior-free discovery system, because the candidate families are supplied. The corrected aggregate sweep was incomplete at submission time, so no identifiability counts are reported.

We also pre-specify an \emph{Evidence Ladder}: Gravity varies visible temporal span at roughly fixed row count; Spring varies amplitude coverage. The hidden mechanism and agent configuration are held fixed. Four ordered evidence levels crossed with five seeds and two families give 40 planned fresh-agent trials. The confirmatory dose-response test is a monotonic trend in mechanism recovery with evidence level, supplemented by logistic regression against $\emargin_{\mathrm{BIC}}(D)$. These runs were not complete at submission time, so no Ladder effect size or trend statistic is reported.

\section{Experimental Protocol}

\paragraph{Agent.}
The benchmark is model-agnostic, but the present empirical checkpoint evaluates one instantiation: a fresh tool-augmented Claude Code sub-agent with Bash, Python, NumPy, SciPy, and pandas, following the broader tool-using/experiment-agent paradigm \citep{schick2023toolformer,huang2024mlagentbench}. The single-agent design is intentional at this stage: it isolates evidence updating before introducing social dynamics from debate, critic, or consensus pipelines. The agent receives a self-contained task and observation table and may create scratch analyses. Ground-truth law metadata, graders, simulator source, and prior results are not placed in the run workspace. Because Bash executes on a shared host, this is \emph{task-level content isolation}, not an OS security sandbox. Raw responses and tool traces are retained for audit.

\paragraph{Pre-specified replication.}
The headline grid crosses six families, five seeds $\{101,202,303,404,505\}$, and famous/twin conditions:
\[
6\times5\times2=60.
\]
The seed indexes deterministic world-generation randomness, including initial-condition sampling and observation perturbations where configured by the generator. Famous and twin trials use the same seed index so that the planned analysis is paired by family and seed. Prompts, generators, observation process, train/holdout/transfer split, predictive threshold, and evaluation logic are frozen for aggregation. A genuine infrastructure failure may be rerun only if no completed scientific answer exists; a completed wrong answer is never retried because of its outcome.

\paragraph{Evaluator development.}
During evaluator development, mathematically valid responses exposed representation-level parser gaps (for example derivative notation, auxiliary definitions, Unicode symbols, and equation separators). These were repaired with generic regression tests while preserving the raw transcripts. One parser version was then used to reprocess the completed transcripts. We report parser failures separately rather than silently converting them into scientific failures. Observation tables retain the benchmark's frozen perturbation and rounding process where applicable; we do not report a separate noise-sensitivity sweep, because evidence quality is treated as a distinct robustness axis rather than folded into the prior-conflict comparison.

\paragraph{Planned primary analysis.}
The confirmatory design compares famous and twin outcomes by matched family--seed pairs. For outcome $Z\in\{\mathrm{predictive},\mathrm{mechanism}\}$, the primary effect is
\[
\Delta_Z
=
\Pr(Z=1\mid \mathrm{Famous})
-
\Pr(Z=1\mid \mathrm{Twin}).
\]
The pre-specified paired analysis uses McNemar's exact test on discordant family--seed pairs and an exact paired permutation interval for $\Delta_Z$. As a secondary model, a mixed-effects logistic regression uses condition as a fixed effect with family and seed random intercepts. Because only two famous controls were completed at this submission checkpoint, none of these confirmatory paired statistics is reported from the partial matrix.

\section{Results}

\subsection{Current replication status}

At submission time, 23/60 headline cells had been attempted: 22 produced graded trial records and one remained a current infrastructure failure; 37 were not yet run. Two of the 22 graded records were unparseable under the frozen evaluator. The completed scientific matrix is highly imbalanced (2 famous, 20 twin), so the rates below are descriptive.

\begin{table}[t]
\caption{\textbf{Submission-time outcome accounting.} Each entry is successes/graded trial records for that cell. Seeds index matched deterministic world-generation randomness, including initial conditions and configured observation perturbations. Parser failures remain visible in the seed matrix; dashes denote cells with no graded record, not zero performance.}
\label{tab:results}
\centering
\small
\begin{tabular}{lcccc}
\toprule
& \multicolumn{2}{c}{Predictive pass} &
\multicolumn{2}{c}{Mechanism recovery}\\
\cmidrule(lr){2-3}\cmidrule(lr){4-5}
Family & Famous & Twin & Famous & Twin\\
\midrule
Spring       & 1/1 & 0/5 & 1/1 & 0/5\\
Gravity      & 1/1 & 1/5 & 1/1 & 4/5\\
Drag         & --  & 5/5 & --  & 1/5\\
Pendulum     & --  & 1/2 & --  & 0/2\\
Conservation & --  & 1/2 & --  & 0/2\\
Coupling     & --  & 0/1 & --  & 0/1\\
\midrule
Overall      & 2/2 & 8/20 & 2/2 & 5/20\\
\bottomrule
\end{tabular}
\end{table}

The raw predictive rates are 100\% (2/2; Wilson 95\% CI $[0.34,1.00]$) in completed famous trials and 40\% (8/20; $[0.22,0.61]$) in twins. Mechanism recovery is 100\% (2/2; $[0.34,1.00]$) versus 25\% (5/20; $[0.11,0.47]$). These are \emph{not} interpreted as a completed matched prior penalty because the famous condition is mostly missing.

\begin{figure}[t]
    \centering
    \includegraphics[width=0.82\linewidth]{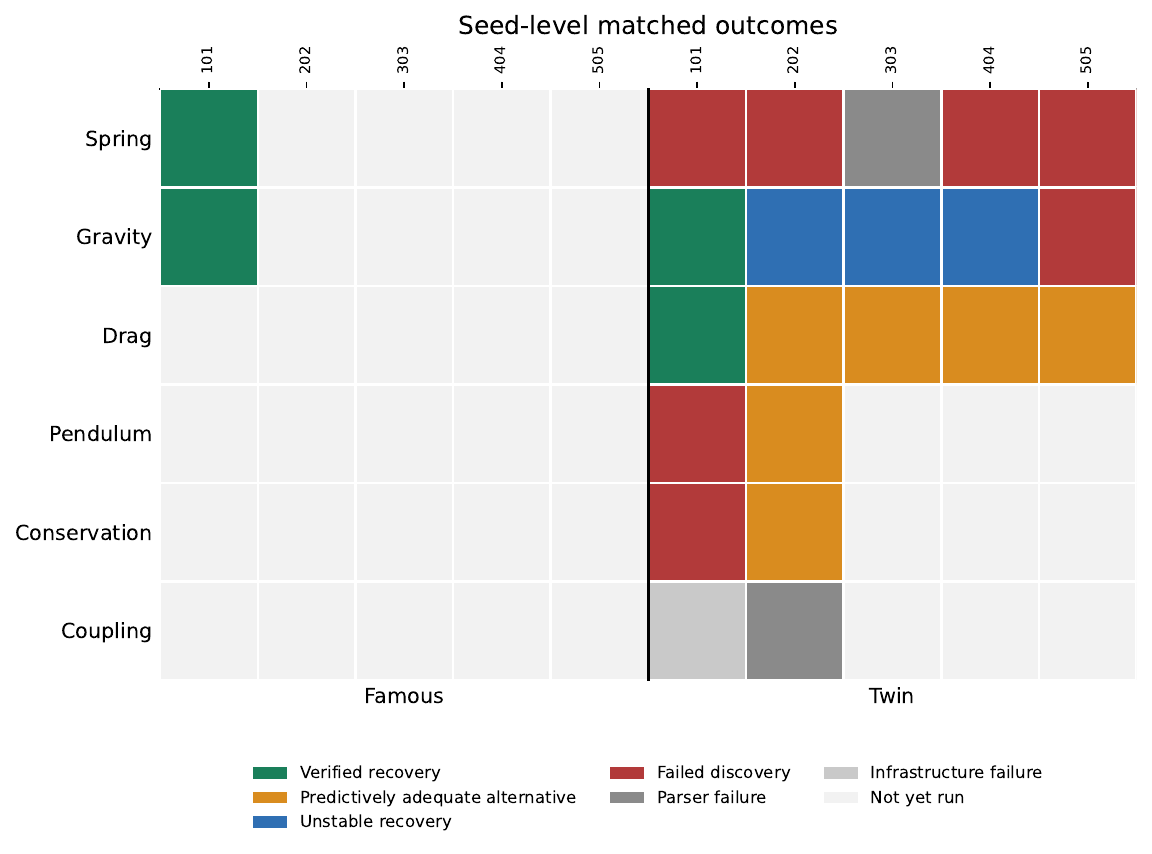}
    \caption{\textbf{Seed-level submission-time status.}
    Missing, parser, and infrastructure cells remain explicit rather than being converted into scientific failures.}
    \label{fig:seedmatrix}
\end{figure}

\subsection{Prediction and mechanism dissociate in both directions}

Among the 20 parsable graded trials, the four-quadrant taxonomy contains 4 verified recoveries, 6 predictively adequate alternatives, 3 unstable recoveries, and 7 failed discoveries; two additional graded responses are reported separately as parser failures. Both off-diagonal quadrants are populated, so a single binary ``success'' score would obscure qualitatively different scientific outcomes.

\begin{figure}[t]
    \centering
    \includegraphics[width=0.58\linewidth]{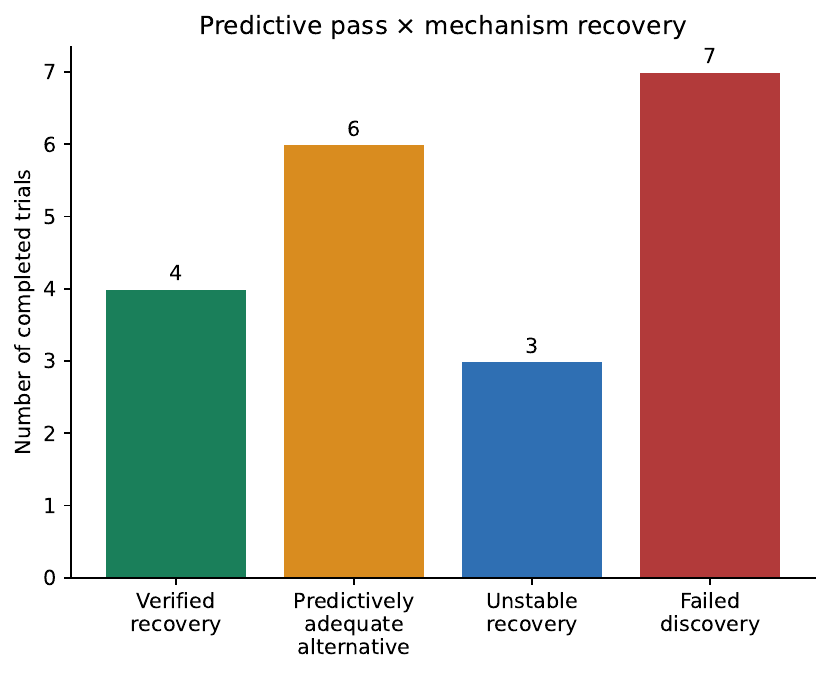}
    \caption{\textbf{Prediction and mechanism recovery dissociate.}
    Counts sum to the 20 parsable graded trials; the two unparseable records from the 22 graded trials are excluded from this four-cell decomposition and remain visible in Fig.~\ref{fig:seedmatrix}.}
    \label{fig:taxonomy}
\end{figure}

\paragraph{Predictive success, wrong mechanism.}
On Drag, all five completed twins pass predictive verification, but only one recovers the noncanonical exponent. In one audited case, the agent's own free-exponent analysis favored a mildly super-linear law near $\hat p\approx1.54$; it nevertheless selected familiar linear drag on simplicity grounds. That reported law still obtained
\[
\nrmse_H=0.0228,\qquad \nrmse_T=0.0840.
\]
A Pendulum twin similarly returned \mbox{$\ddot{\theta}=-1.539\,\theta$} rather than the nonlinear generator while obtaining $\nrmse_H=0.0016$ and $\nrmse_T=0.0068$. An unseen test certifies only distinctions that the test actually exposes.

\paragraph{Mechanism recovery, predictive failure.}
Gravity exhibits the opposite pattern. Across five twisted-gravity seeds, the mechanism checker recovers the non-Newtonian exponent near $p=2.3$ in 4/5 cases, but only 1/5 passes the long-horizon predictive verifier. Three trials are therefore \emph{unstable recoveries}. In an audited run, the agent explicitly used apsidal precession to reject $p=2$ and recover $\hat p\approx2.30$. The remaining rollout failures are not relabeled: small coefficient errors accumulating over multi-orbit integration are a plausible post-hoc explanation, not part of the mechanism criterion.


\subsection{Benchmark-audit findings}
\label{sec:audit}

The experiment also exposed benchmark-design failures that should not be hidden. In the Conservation family, independent twin agents repeatedly discovered the energy-like invariant \mbox{$Q(s,v)=s\,v^2$} to the numerical noise floor, indicating a shared or degenerate invariant that defeats the intended structural contrast. We therefore retain those trials in end-to-end accounting but do not use Conservation as evidence for a prior-conflict mechanism. Coupling currently has only one completed twin and remains too sparse for a family-level claim. Before either structural family returns to the headline analysis, the generator must pass the non-degeneracy checks in Appendix~\ref{app:structural}, including algebraic-independence tests over the sampled evaluation manifold and transfer rejection of trivial separable invariants. The strongest scientific interpretation of the current checkpoint therefore comes from the audited Spring, Gravity, Drag, and Pendulum families.

\section{Discussion}

The first lesson is a verification one:
\[
\boxed{\text{predictive adequacy}\neq\text{mechanism recovery}.}
\]
A model can tell the wrong scientific story and still survive a held-out test. The reverse also occurs: it can identify the right mechanism and execute it badly enough to fail a rollout. Those are not cosmetic distinctions. A scientific verifier should say exactly what it certifies: prediction under specified interventions, recovery of a mechanism, or both, rather than collapsing them into a single success bit \citep{popper1959logic,mossel2025refutability}. The Gravity and Drag contrast also exposes a metric asymmetry: pointwise long-horizon error is phase-sensitive in conservative dynamics, while dissipative dynamics can make structurally different models converge to similar trajectories. This is why we preserve the frozen rollout label but separate it from mechanism recovery and pre-specify complementary diagnostics rather than silently changing the threshold after seeing outcomes.

Second, verification is partly an experimental-design problem. If competing mechanisms make nearly identical predictions under the available intervention, a holdout set cannot distinguish them merely by being unseen. This motivates the hypothesis-family-aware identifiability control and the Evidence Ladder: the relevant question is not only whether the agent failed, but whether the experiment supplied discriminative evidence.

Third, scientific priors are useful inductive bias rather than a defect. Equation-discovery systems exploit priors to search huge hypothesis spaces efficiently. \systemname{} studies the complementary failure mode: when a familiar prior is wrong, how much evidence is required to escape it? The framework is not intrinsically tied to ODEs: in probabilistic or event-based sciences, NRMSE can be replaced by likelihood/proper-scoring verification and transfer can be expressed through held-out interventions or invariance tests \citep{peters2016invariant}. Structured observational programs such as cliodynamics provide a natural non-physics analogue of theory discrimination from longitudinal evidence \citep{turchin2003historical,turchin2015seshat,turchin2018complexity}.

\section{Limitations and Pre-Specified Follow-ups}

The largest limitation is incompleteness: 23/60 headline cells had been attempted and only two famous controls were complete. The current numbers therefore demonstrate the benchmark and the prediction--mechanism dissociation, not the magnitude of a causal prior-conflict penalty. The immediate confirmatory work is to complete the matched grid, the complexity-aware identifiability sweep, and the 40-run Evidence Ladder, then report the pre-specified McNemar, paired-permutation, mixed-effects, and ordered-trend analyses. A controlled noise sweep is also needed to separate evidence quality from prior conflict.

Generality remains open. Prior-free symbolic baselines such as PySR, SINDy, and AI Feynman, cross-model replication, LLM-guided symbolic search, and an explicit critic/falsifier condition would test whether the observed behavior is specific to the present agent architecture \citep{brunton2016sindy,udrescu2020feynman,cranmer2023pysr,shojaee2025llmsr,bansal2026pcf}. External evaluation on LLM-SRBench and NewtonBench would test whether the two-axis taxonomy transfers beyond our generators \citep{shojaee2025llmsrbench,zheng2025newtonbench}. Conservation and Coupling remain visible in end-to-end accounting but are excluded from strong mechanistic conclusions until they pass the structural non-degeneracy audit and are rerun across all five seeds. Visible-analysis truncation remains a behavioral pilot only because visible chain-of-thought need not faithfully expose hidden computation \citep{turpin2023unfaithful,lanham2023faithfulness}.

\section{Conclusion}

\systemname{} asks a stricter question than whether an AI scientist can produce a plausible equation: \emph{what did the evidence actually verify?} In the completed subset, Gravity often identifies the changed mechanism without stable rollout, while Drag often predicts successfully with the wrong mechanism. The incomplete matched matrix prevents a confirmatory estimate of prior conflict, but the completed runs already show why prediction alone is insufficient. Scientific verification should distinguish whether a law predicts under specified interventions, recovers the mechanism supported by the evidence, or does both.

\bibliographystyle{plainnat}
\bibliography{synthetic_universes_refs}

\clearpage
\appendix

\raggedbottom
\setlength{\textfloatsep}{9pt plus 2pt minus 2pt}
\setlength{\floatsep}{8pt plus 2pt minus 2pt}
\setlength{\intextsep}{8pt plus 2pt minus 2pt}
\renewcommand{\topfraction}{0.92}
\renewcommand{\bottomfraction}{0.75}
\renewcommand{\textfraction}{0.05}
\renewcommand{\floatpagefraction}{0.80}


\section{Audit Trail, Reproducibility, and Run Integrity}
\label{app:repro}

The benchmark is intended to make scientific failure inspectable rather than merely scoreable. Each attempted trial produces a persistent record containing the task, observation table, raw agent response, parsed answer when available, predictive-verifier outputs, mechanism-recovery output, and infrastructure status. These artifacts are retained even when a run fails. This matters because an autonomous-science benchmark can otherwise improve its apparent performance simply by losing difficult runs to parser, API, or execution failures.

\subsection{What the agent can and cannot see}

Each scientific trial starts from a fresh task workspace containing only the natural-language task, the observation table, and an empty scratch directory. The agent may use Bash and scientific Python tools to inspect the evidence, fit models, integrate candidate dynamics, and create intermediate analyses. Ground-truth generator metadata, grader code, simulator source, and previous result files are excluded from that workspace. Bash still runs on a shared host, so we describe this as \emph{task-level content isolation}, not as an operating-system security sandbox. Raw tool traces are retained so that out-of-workspace accesses can be audited rather than assumed away.

This distinction is deliberate. The experimental claim is about behavior under a controlled information interface; it is not a claim that the execution environment constitutes a hardened security boundary. A stronger release should reproduce the same trials in containerized sandboxes and compare the resulting labels.

\subsection{Failure accounting}

We separate three failure modes that are often conflated:
\begin{enumerate}
    \item \textbf{Scientific failure:} the agent produces a completed, parsable law that fails predictive verification, mechanism recovery, or both.
    \item \textbf{Representation failure:} the agent completes the task but the frozen parser cannot turn the reported law into an executable representation.
    \item \textbf{Infrastructure failure:} the run terminates before a scientific answer exists because of an API/session/runtime failure.
\end{enumerate}
Only the first category is evidence about scientific reasoning. Parser and infrastructure failures remain part of end-to-end system reliability, which is why they are displayed explicitly in Fig.~\ref{fig:seedmatrix} instead of being silently discarded.

\subsection{Evaluator development and freezing}

During evaluator development, real completed responses exposed notation gaps such as Unicode superscripts, Leibniz derivative notation, auxiliary definitions, and explanatory prose attached to equations. Repairs were made at the representation layer and regression-tested against preserved transcripts. The evaluator was then re-run uniformly over the completed set. No scientific answer was retried because it was wrong, and mechanism labels are computed independently of predictive pass.

The release manifest should include the exact model snapshot and generation settings; prompt text; seeds; observation and perturbation settings; train/holdout/transfer ranges; Python/package versions; ODE solver and tolerances; timeouts and interaction budget; raw responses and tool traces; parser version; predictive grader outputs; mechanism-checker source; and infrastructure-failure logs. For the identifiability control it should additionally include the definitions of $\mathcal F_F$ and $\mathcal F_T$, nesting relation, loss $E(f;D)$, parameter bounds, deterministic initialization grid, restart count, optimizer, stopping tolerances, and the exact value of $\epsilon$.

\begin{table}[!htbp]
\renewcommand{\arraystretch}{1.08}
\caption{\textbf{Release artifacts and what they make auditable.} The point of the release is not only rerunning a headline score, but reconstructing why a trial received its label.}
\label{tab:release}
\centering
\small
\begin{tabularx}{\linewidth}{lX}
\toprule
Artifact & Audit question\\
\midrule
Task + observations & What evidence was actually available to the agent?\\
Raw response + tool trace & Which hypotheses and calculations were externally visible?\\
Predictive grader output & Did the reported law pass continuation and transfer?\\
Mechanism-checker output & Did the executable law match the generating equivalence class?\\
Parser regression tests & Could notation handling change a scientific label?\\
Failure logs & Was a missing result scientific, representational, or infrastructural?\\
\bottomrule
\end{tabularx}
\end{table}

\section{Mechanism Recovery Is an Executable Criterion}
\label{app:mechanism}

Equation~\ref{eq:mechanism} is intentionally stricter than textual equation matching. A response can be algebraically different but scientifically equivalent, or textually similar while encoding the wrong response function. The evaluator therefore parses the answer into an executable object and compares behavior on a fixed domain. Continuous families recover a response-implied shape parameter after accounting for nuisance scale; structural families test an equivalence class.

\begin{table}[!htbp]
\renewcommand{\arraystretch}{1.08}
\caption{\textbf{Mechanism quantities used by the frozen checker.} The functional residual gate is $0.05$ where applicable; the released source contains the fixed family-specific shape gates and evaluation domains.}
\label{tab:mechrules}
\centering
\small
\begin{tabularx}{\linewidth}{lXX}
\toprule
Family & Scientific distinction & Audit quantity\\
\midrule
Spring & restoring response $p=1$ vs.\ $p=1.3$ & matched power $p$; functional residual\\
Gravity & central-force exponent $p=2$ vs.\ $p=2.3$ & matched exponent $p$; functional residual\\
Drag & velocity exponent $p=1$ vs.\ $p=1.6$ & matched exponent $p$; functional residual\\
Pendulum & sinusoidal response vs.\ amplitude-dependent correction & fixed-domain functional residual / shape fit\\
Conservation & invariant equivalence class & normalized structural/equivalence residual\\
Coupling & pairwise interaction structure & product-vs.-alternative structural check\\
\bottomrule
\end{tabularx}
\end{table}

\subsection{Audited examples that motivate the two-axis verifier}

The completed trials make the need for an executable mechanism check concrete.

\paragraph{Drag: a good forecast from the wrong law.}
One twisted Drag run selected familiar linear drag even after its own free-exponent analysis admitted a mildly super-linear alternative. The reported law nevertheless achieved $\nrmse_H=0.0228$ and $\nrmse_T=0.0840$, so a prediction-only benchmark would certify it. The mechanism checker instead matches a velocity exponent near $p=1.0$ against the twisted generator's $p=1.6$ and rejects mechanism recovery. Across the five completed Drag twins, predictive verification is $5/5$ while mechanism recovery is $1/5$.

\paragraph{Gravity: the right exponent with an unstable rollout.}
The opposite failure appears in Gravity. Four of five twisted-gravity runs recover the non-Newtonian exponent near $p=2.3$, but only one passes both rollout tests. In one audited transcript the agent uses apsidal precession, rather than a memorized label, to reject $p=2$ and estimate $p\approx2.30$. These cases are counted as \emph{unstable recoveries}: the scientific mechanism is identified, but execution is not accurate enough to satisfy the external predictive verifier.

\paragraph{Pendulum: why unseen data can still be non-discriminative.}
A twisted Pendulum run reports the linear oscillator $\ddot\theta=-1.539\theta$ and achieves $\nrmse_H=0.0016$ and $\nrmse_T=0.0068$. The functional checker, evaluated on a wider fixed angular domain, rejects the nonlinear mechanism match. This is the core experimental-design warning of the paper: an unseen test is not automatically a discriminative test.

Figure~\ref{fig:familymech} gives the family-level diagnostic view. It is kept in the appendix because Table~\ref{tab:results} already carries the numerical result in the main paper, while the plot is useful for seeing the opposite Gravity--Drag dissociation at a glance.

\IfFileExists{fig_mechanism_recovery_by_family_final.pdf}{
\begin{figure}[!htbp]
\centering
\includegraphics[width=0.68\linewidth]{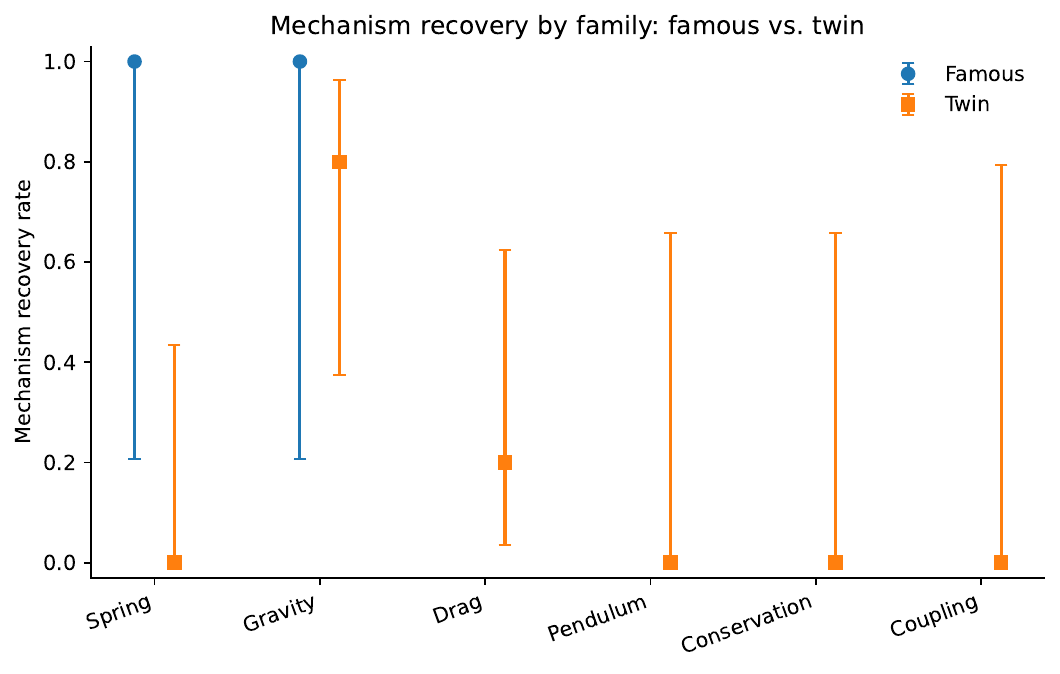}
\caption{\textbf{Family-level mechanism recovery in the current checkpoint.}
The figure is descriptive because famous controls are mostly missing. Its main value is diagnostic:
Gravity and Drag exhibit opposite relationships between predictive success and mechanism recovery,
which is why the paper reports the two criteria separately rather than averaging them into one discovery score.}
\label{fig:familymech}
\end{figure}}{}

\section{Identifiability: Was the Experiment Capable of Distinguishing the Law?}
\label{app:ident}

A failed agent is not evidence of a stubborn prior if the observations themselves do not distinguish the competing hypotheses. The hypothesis-family-aware control gives a numerical optimizer more structural information than the language agent receives. It fits the canonical and twin-aware families using the normalized trajectory loss in Eq.~\ref{eq:identloss}, then compares them with the complexity-adjusted margin $\emargin_{\mathrm{BIC}}(D)$ in Eq.~\ref{eq:bicmargin}. This correction matters because the continuous canonical family is nested inside the more flexible twin-aware family; raw training residual alone would mechanically favor the larger family.

This baseline should not be confused with symbolic discovery. PySR, SINDy, AI Feynman, and hybrid LLM-guided symbolic search operate over substantially larger hypothesis spaces \citep{brunton2016sindy,udrescu2020feynman,cranmer2023pysr,shojaee2025llmsr}. The oracle control answers a narrower question: if even a supplied twin-aware family cannot recover a model that survives external holdout and transfer verification, the instance is underidentified and should not be used as clean evidence of prior interference.

The corrected full identifiability sweep was not complete at submission time, so we intentionally omit aggregate oracle-identifiability counts. The final table should report $E_F(D)$, $E_T(D)$, model dimensions $k_F,k_T$, $\mathrm{BIC}_F$, $\mathrm{BIC}_T$, $\emargin_{\mathrm{BIC}}(D)$, a visible-data cross-validation check, the oracle model's holdout/transfer result, and the agent mechanism label on the same row.

\IfFileExists{figure_verification_stages.png}{
\begin{figure}[!htbp]
\centering
\includegraphics[width=0.64\linewidth]{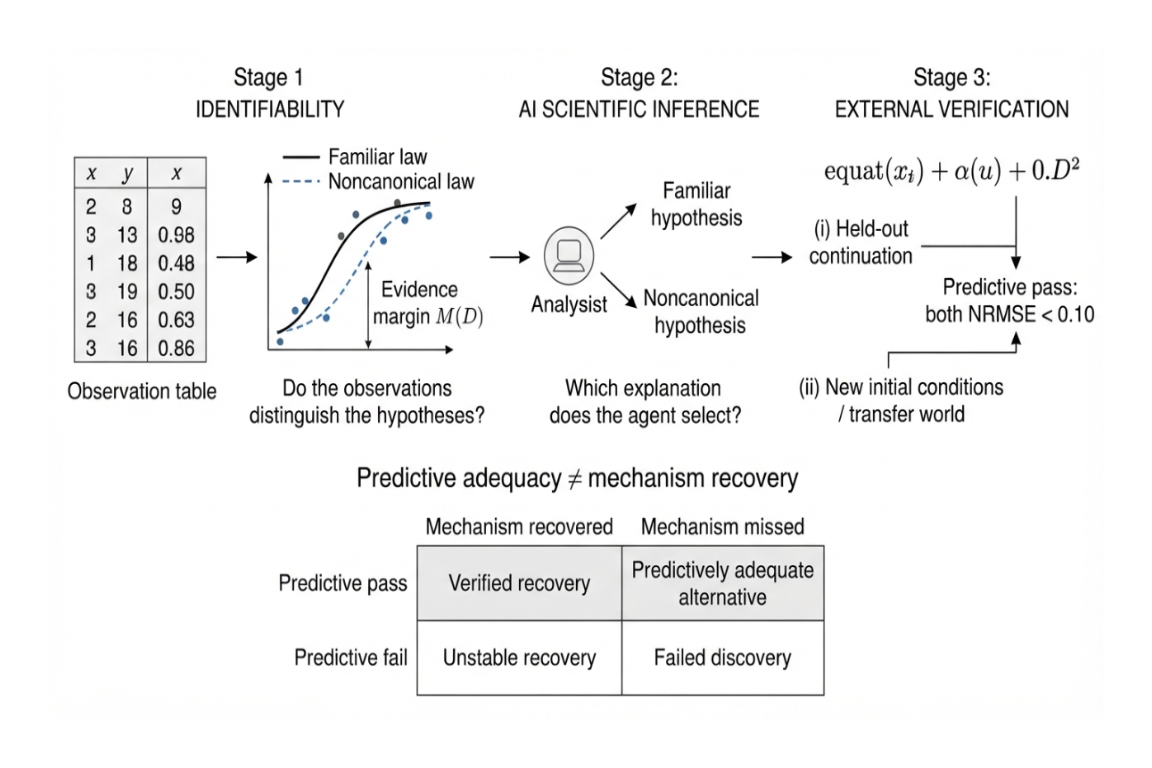}
\caption{\textbf{Failure localization used by \systemname{}.}
Stage 1 asks whether the available evidence discriminates candidate families; Stage 2 asks which law the agent actually reports; Stage 3 executes that law externally. The four-quadrant outcome is therefore the end of a diagnostic chain, not the whole diagnosis.}
\label{fig:stages}
\end{figure}}{}

\section{Evidence Ladder: A Pre-Specified Falsification Experiment}
\label{app:ladder}

Cross-family comparisons cannot establish that evidence strength \emph{causes} prior abandonment: Gravity, Drag, Spring, and Pendulum differ in dynamics as well as in evidentiary geometry. The Evidence Ladder was therefore designed as a within-family intervention. It changes the observations while keeping the hidden law fixed.

For Gravity, the intervention increases visible temporal/orbital span at approximately fixed row count, exposing progressively more apsidal structure. For Spring, it increases amplitude coverage, exposing progressively more of the nonlinear response. Four ordered conditions crossed with five seeds in each family yield 40 fresh-agent trials. The model, hidden law, tool access, and evaluation code remain fixed.

Figure~\ref{fig:ladder} visualizes the intervention. The quantity being increased is not simply sample count; it is coverage of the part of state space where the familiar and twisted mechanisms make meaningfully different predictions.

\IfFileExists{figure_evidence_ladder.png}{
\begin{figure}[!htbp]
\centering
\includegraphics[width=0.67\linewidth]{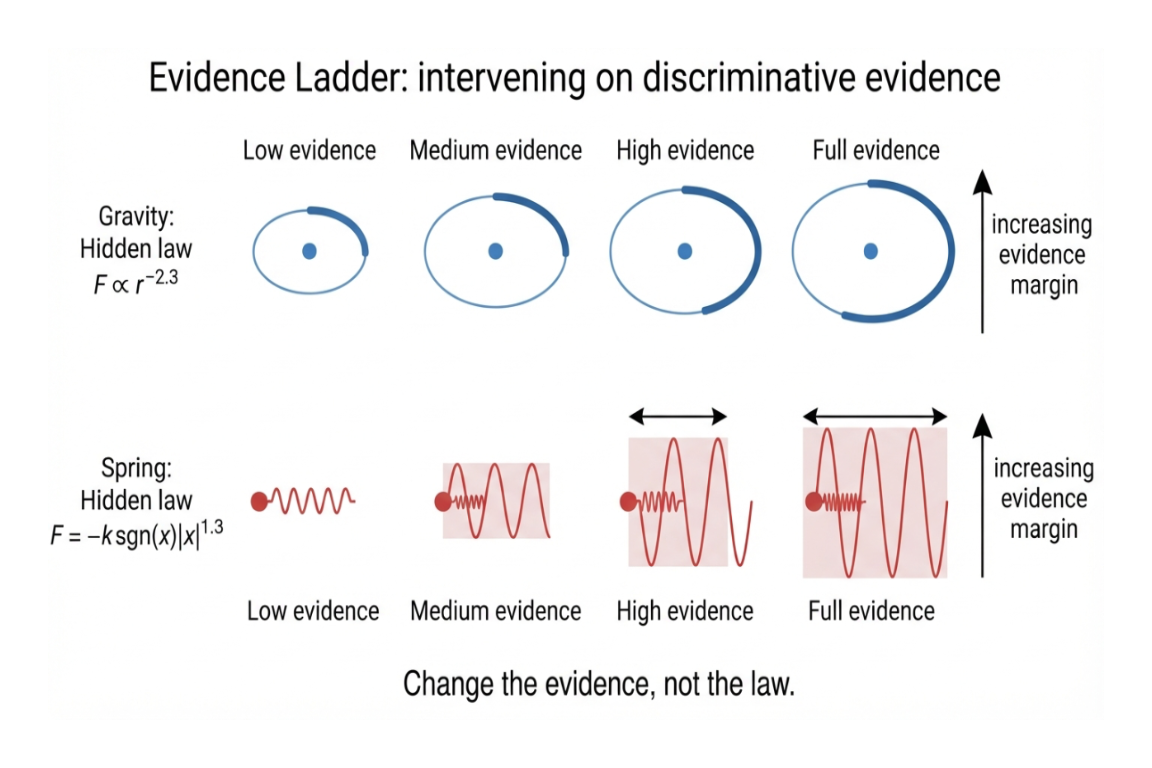}
\caption{\textbf{Evidence Ladder design: change the evidence, not the law.}
The intervention is intended to move the same hidden mechanism from weakly discriminative to strongly discriminative observations.
A completed positive result would require recovery to increase with evidence strength; the schematic itself is not evidence for that trend.}
\label{fig:ladder}
\end{figure}}{}

The Ladder is the cleanest falsification test of a causal ``stronger evidence overcomes a familiar prior'' interpretation. The primary trend analysis is a monotonic test across the four ordered evidence levels, with logistic regression of mechanism recovery on $\emargin_{\mathrm{BIC}}(D)$ as a continuous secondary analysis. Three outcomes would weaken the interpretation: (i) recovery fails to increase across ordered evidence levels; (ii) the oracle evidence margin increases but agent recovery does not; or (iii) recovery changes without a corresponding change in discriminative evidence. Because the full sweep is incomplete, none of these causal alternatives is resolved in the present submission.

\section{Dynamical Sensitivity of the Predictive Verifier}
\label{app:dynamics}

The headline predictive label deliberately remains the frozen dual-threshold test in Eq.~\ref{eq:verify}; changing it after observing Gravity and Drag would compromise comparability. The reviewer concern about dynamical regime sensitivity is nevertheless substantive, so we pre-specify three secondary diagnostics that localize why a rollout failed without changing the original label.

\paragraph{Raw versus calibrated structural rollout.}
The raw rollout uses the agent's reported structure and numerical parameters exactly as stated. A secondary calibrated rollout keeps the parsed functional form fixed but refits only nuisance scale parameters by least squares on visible training data, then evaluates the same holdout and transfer trajectories. A large improvement after calibration indicates that the mechanism was structurally useful but numerically miscalibrated.

\paragraph{Derivative-field error.}
For dynamical laws with state $\mathbf y$ and derivative field $g_f(\mathbf y)$, define
\[
\nrmse_{\dot Y}
=
\frac{
\sqrt{\frac{1}{N}\sum_{i=1}^{N}
\|g_{\hat f}(\mathbf y_i)-g_{f^\star}(\mathbf y_i)\|_2^2}
}{
s_{\dot Y}
}.
\]
This removes cumulative integration phase drift and tests local dynamical agreement directly.

\paragraph{Geometry-aware diagnostics.}
For periodic and orbital systems, we additionally report family-appropriate quantities such as frequency error, apsidal-precession error, invariant drift, and rollout error over standardized fractions of a characteristic period. These metrics are diagnostics, not replacement headline outcomes. Their purpose is to distinguish wrong dynamics from correct dynamics whose long-horizon phase is numerically unstable.

\section{Pilot Analyses: What They Contributed, and Why They Are Not Headline Evidence}
\label{app:pilots}

The pilot stage was used to discover benchmark failure modes, stress the parser, and decide which comparisons deserved replication. It is separated from the frozen replication because pilot observations influenced subsequent analysis design.

\subsection{First-pass outcomes generated the replication hypotheses}

Figure~\ref{fig:firstpass} is retained as provenance for the replication design. It records the pilot pattern that motivated the frozen taxonomy, but it is not pooled into the submission-time replication denominator.

\IfFileExists{fig_firstpass_outcomes.png}{
\begin{figure}[!htbp]
\centering
\includegraphics[width=0.60\linewidth]{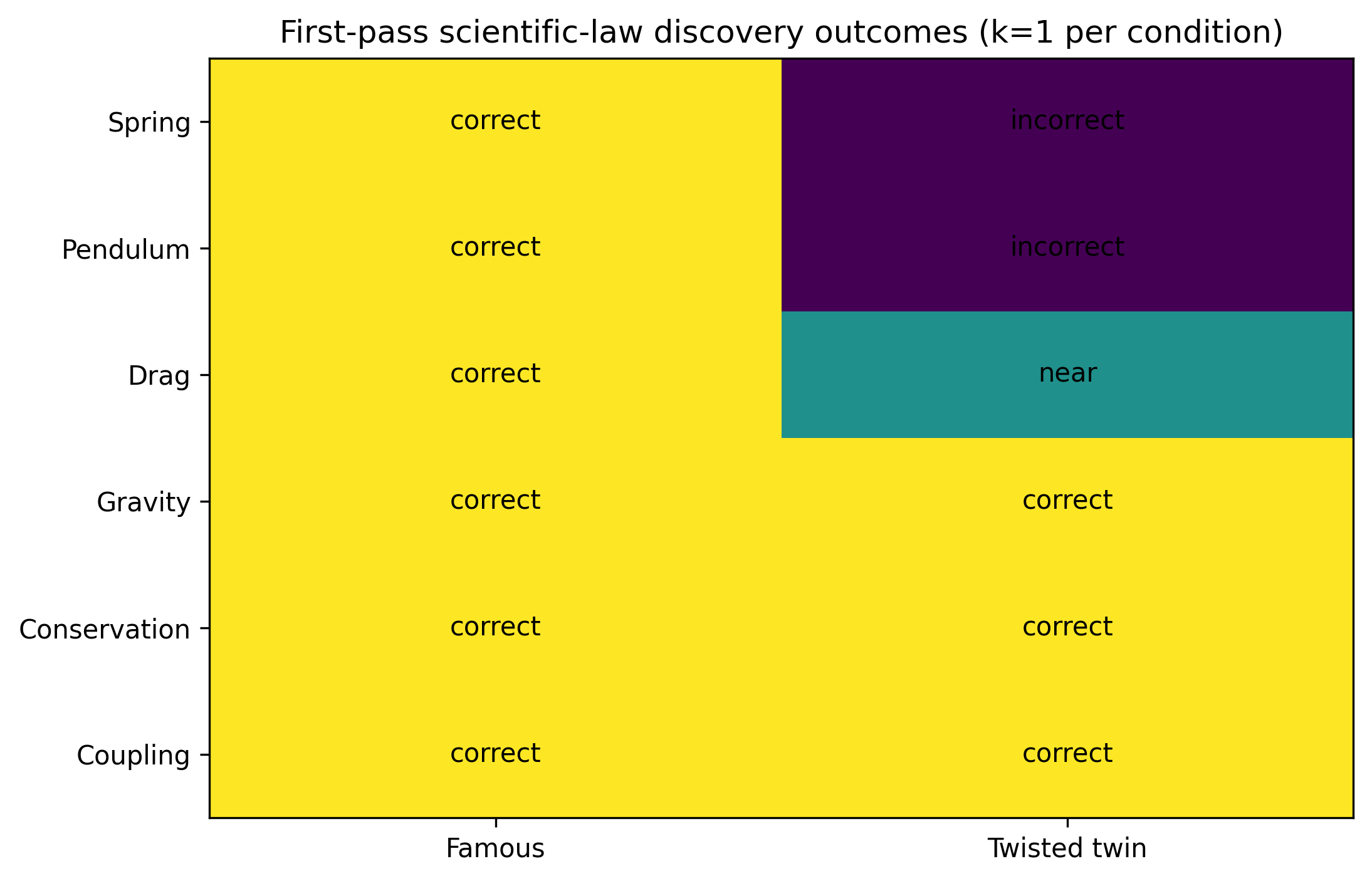}
\caption{\textbf{First-pass outcomes ($k=1$ per condition).}
These results motivated the replicated matrix and the prediction--mechanism taxonomy. They are shown for provenance, not pooled with the submission-time replicated counts.}
\label{fig:firstpass}
\end{figure}}{}

The pilot suggested two qualitatively different failure modes. Some twins appeared to preserve predictive performance despite a familiar or near-familiar reported law; others appeared to drive the agent toward the noncanonical mechanism. Those observations motivated the frozen four-cell taxonomy and the decision to track mechanism recovery separately from predictive pass. Once the replication began, pilot results were not used as additional samples.

\subsection{Visible-prefix truncation is a behavioral probe, not a window into hidden reasoning}

The first-pass ``Scissors'' probe truncates the \emph{visible} analysis preceding the final reported law. It asks a descriptive question: at what retained prefix does the reported conclusion change? It does not provide privileged access to hidden computation, and visible chain-of-thought can be an unfaithful account of the computation producing an answer \citep{turpin2023unfaithful,lanham2023faithfulness}.

The visible-prefix result in Fig.~\ref{fig:scissors} is kept only as a behavioral audit. It shows that the reported law can be sensitive to which externally visible analysis survives, without treating visible chain-of-thought as privileged access to the model's hidden computation.

\IfFileExists{fig_scissors_gravity_firstpass.png}{
\begin{figure}[!htbp]
\centering
\includegraphics[width=0.66\linewidth]{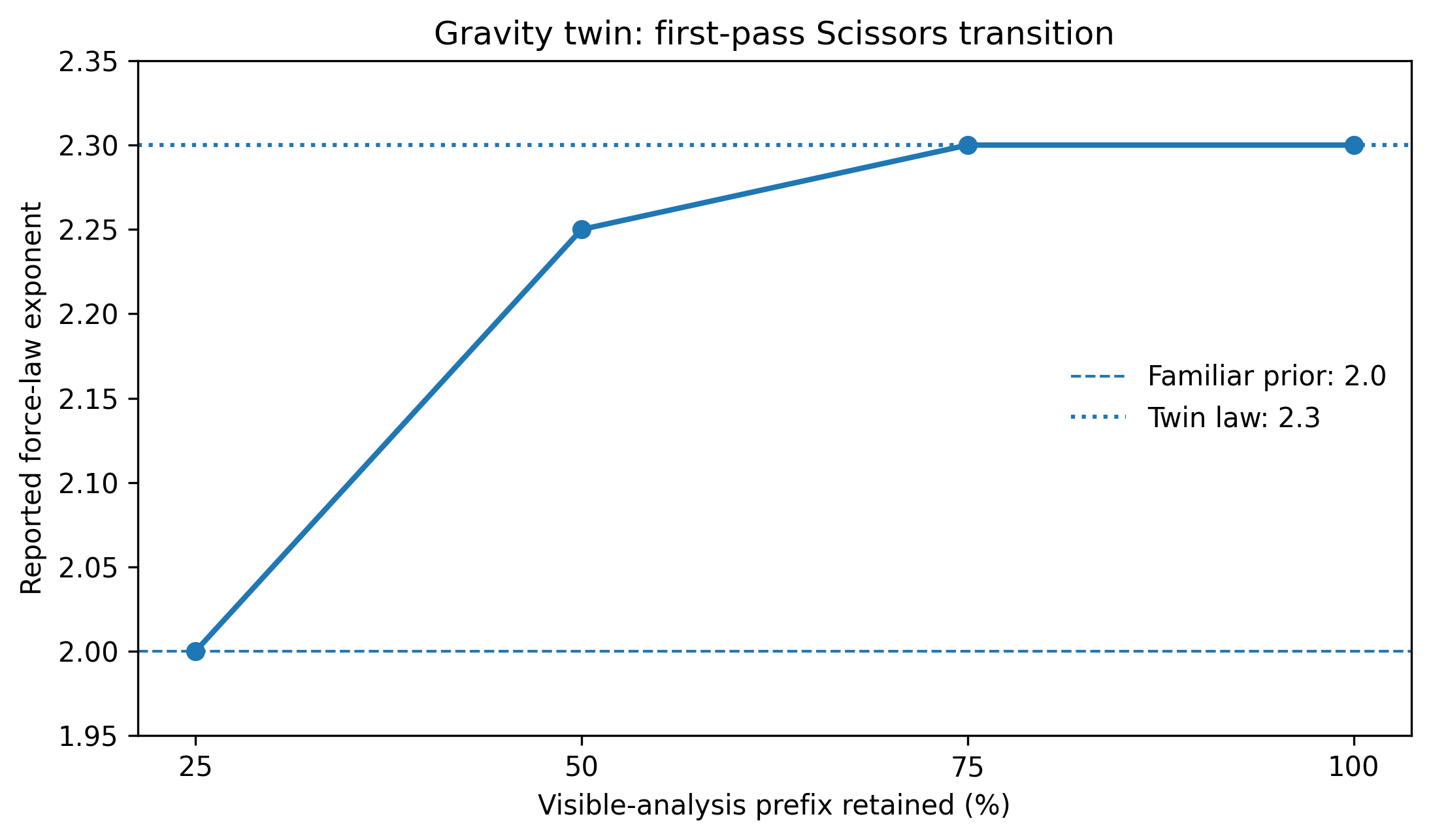}
\caption{\textbf{First-pass Gravity visible-prefix transition.}
The reported exponent moves from the familiar value toward the twisted law as more visible analysis is retained.
Because this is a single pilot trajectory and operates on visible text rather than hidden reasoning, it is not used to support the paper's causal claims.}
\label{fig:scissors}
\end{figure}}{}

The plot is still useful as an audit artifact: it shows that the final law is sensitive to which externally visible analysis survives. But the stronger experiment is the Evidence Ladder, which intervenes on the scientific evidence itself rather than on a textual trace. This is why the main paper treats Scissors as provenance and the Ladder as the pre-specified causal test.

\section{Structural-Family Non-Degeneracy Audit}
\label{app:structural}

The Conservation failure shows that a structural benchmark must verify the task itself before evaluating the agent. We therefore make the acceptance criteria for future structural families explicit rather than retroactively pretending the current generator satisfies them.

Let $Q_1,\ldots,Q_m$ be the intended invariant basis over state $\mathbf x$. On a sampled evaluation manifold $\mathcal X$, define the stacked Jacobian
\[
J_Q(\mathbf x)
=
\begin{bmatrix}
\nabla Q_1(\mathbf x)^\top\\
\vdots\\
\nabla Q_m(\mathbf x)^\top
\end{bmatrix}.
\]
A structural family is accepted only if the intended invariant basis is locally non-degenerate over the evaluation support, meaning $J_Q$ has the intended rank away from known symmetry or singular sets after quotienting irrelevant global scale. In addition, a library of low-order separable and kinematic candidates must fail conservation on independently generated transfer events whenever those candidates are not part of the target equivalence class.

For coupling tasks, the corresponding audit requires genuine cross-variable dependence: the target interaction must not decompose into independent single-body terms over the sampled support, and alternative separable forms must be rejected on transfer cases. These checks are design-time gates. Conservation and Coupling will not return to the headline mechanism analysis until they pass them and are rerun across all five pre-specified seeds.

The current submission therefore reports the observed Conservation degeneracy, \mbox{$Q(s,v)=s\,v^2$}, as a benchmark-audit finding rather than as evidence about model priors. This is a stricter interpretation than substituting the originally intended equations after the fact.

\section{Claim Ledger and Falsifiers}
\label{app:claims}

The purpose of this appendix is to make the paper's epistemic boundary explicit. The strongest claims are those already supported by completed, auditable trials; causal claims about prior conflict are deliberately deferred.

\begin{table}[!htbp]
\renewcommand{\arraystretch}{1.10}
\caption{\textbf{What the current submission claims, and what would overturn it.}}
\label{tab:claims}
\centering
\small
\begin{tabularx}{\linewidth}{p{0.25\linewidth}p{0.34\linewidth}X}
\toprule
Claim & Current support & What would weaken or falsify it\\
\midrule
Predictive adequacy and mechanism recovery are distinct verification targets &
Both off-diagonal quadrants are populated; Drag and Gravity show replicated opposite dissociations. &
A corrected evaluator that makes the off-diagonal cases disappear, or evidence that their labels arise from parser/grader artifacts.\\
\addlinespace
Paired famous/twin worlds provide a controlled way to study prior--evidence conflict &
The observation interface is shared while the hidden mechanism changes. &
A demonstrable cue that reveals world identity, or contamination/leakage that gives access to generator labels.\\
\addlinespace
The present data establish the magnitude of a prior-conflict penalty &
\textbf{Not claimed.} Famous controls are incomplete. &
Requires completion of the matched grid and the pre-specified paired analysis.\\
\addlinespace
Increasing discriminative evidence causes prior abandonment &
\textbf{Not claimed.} The Evidence Ladder is incomplete. &
A flat or reversed recovery trend under increasing complexity-adjusted oracle evidence would directly challenge this interpretation.\\
\addlinespace
The observed twin failures are specific to pretrained language-model priors &
\textbf{Not claimed.} No prior-free or cross-model baseline is complete. &
A symbolic baseline showing comparable failures, or strong dependence on agent architecture, would weaken a pretraining-prior explanation.\\
\addlinespace
The uniform rollout threshold fully characterizes scientific correctness &
\textbf{Not claimed.} Rollout and mechanism are deliberately separate. &
Large agreement on derivative/geometric diagnostics despite raw rollout failure would show that pointwise long-horizon NRMSE is too coarse as a standalone verifier.\\
\bottomrule
\end{tabularx}
\end{table}

This ledger is intentionally conservative. The paper's current empirical contribution is the benchmark design, the execution-grounded two-axis verifier, the transparent audit of benchmark/evaluator failures, and the observed prediction--mechanism dissociation. The unfinished controls determine whether that dissociation can additionally be attributed quantitatively to prior--evidence conflict.

\end{document}